\documentclass[journal]{IEEEtai}

\usepackage[colorlinks,urlcolor=blue,linkcolor=blue,citecolor=blue]{hyperref}

\usepackage{color,array}

\usepackage{graphicx}
\usepackage[nosumlimits]{amsmath}

\begin{document}

\title{Deriving Equivalent Symbol-Based Decision Models from Feedforward Neural Networks}

\author{Sebastian Seidel and Uwe M. Borghoff
\thanks{This work was supported in part by the project \textsc{ROLORAN} within dtec.bw (Digitalization and Technology Research Center of the Bundeswehr). dtec.bw is funded by the European Union's NextGenerationEU program.'}
\thanks{Sebastian Seidel is with the KNDS Deutschland GmbH \& Co.\ KG, 80997 Munich, Germany (e-mail: sebastian.seidel@knds.de).}
\thanks{Uwe M. Borghoff is with the Institute for Software Technology at the University of the Bundeswehr Munich, 85577 Neubiberg, Germany (e-mail: uwe.borghoff@unibw.de).}
}

\markboth{Sebastian Seidel and Uwe M. Borghoff}
{Deriving Equivalent Symbol-Based Decision Models from Feedforward Neural Networks}

\maketitle
\thispagestyle{empty}

\begin{abstract}
Artificial intelligence (AI) has emerged as a transformative force across industries, driven by advances in deep learning and natural language processing, and fueled by large-scale data and computing resources. Despite its rapid adoption, the opacity of AI systems poses significant challenges to trust and acceptance. 

This work explores the intersection of connectionist and symbolic approaches to artificial intelligence, focusing on the derivation of interpretable symbolic models, such as decision trees, from feedforward neural networks (FNNs). Decision trees provide a transparent framework for elucidating the operations of neural networks while preserving their functionality. The derivation is presented in a step-by-step approach and illustrated with several examples. A systematic methodology is proposed to bridge neural and symbolic paradigms by exploiting distributed representations in FNNs to identify symbolic components, including fillers, roles, and their interrelationships. The process traces neuron activation values and input configurations across network layers, mapping activations and their underlying inputs to decision tree edges. The resulting symbolic structures effectively capture FNN decision processes and enable scalability to deeper networks through iterative refinement of subpaths for each hidden layer.

To validate the theoretical framework, a prototype was developed using 
\textsc{Keras} \texttt{.h5}-data and emulating \textsc{TensorFlow} within the \textsc{Java JDK}/\textsc{JavaFX} environment. This prototype demonstrates the feasibility of extracting symbolic representations from neural networks, enhancing trust in AI systems, and promoting accountability.
\end{abstract}

\begin{IEEEImpStatement}
This work addresses the pressing need for confidence and transparency in AI systems by bridging connectionist and symbolic approaches. By providing a clear methodology to demystify the ``black box" nature of AI, it empowers stakeholders---users, developers, and regulators---to better understand and trust AI decisions and the ``rational" actions they produce. The research advances Explainable AI (XAI) by contributing both theoretical insights and practical tools for interpretability, facilitating ethical and effective AI deployment.
This innovation is particularly impactful in sensitive fields such as healthcare, finance, and autonomous systems, where transparency is critical to ensuring accountability and public trust. By fostering greater acceptance of AI technologies, this work accelerates responsible AI integration, shaping a future where innovation aligns with societal needs for ethical decision-making and reliable systems.
\end{IEEEImpStatement}


\section{Introduction}
AI has become a focal point of attention, often described as experiencing a ``hype." This surge of interest is driven by its transformative potential across industries, promising significant efficiency gains and driving innovation, particularly in the areas of image recognition and analysis and natural language processing. The proliferation of large datasets, coupled with increasingly powerful computational resources, has further accelerated the development and deployment of AI applications. As a result, AI has become a cornerstone of modern IT systems, embedded in everything from consumer products to large-scale industrial processes. Despite these advancements, a significant gap remains in terms of trust \cite{ChamolaHSGDS23} and acceptance for AI systems, particularly those perceived as ``black boxes" \cite{AdadiB18}.

This lack of trust has led to a growing interest in XAI. The core goal of XAI is to make AI systems more transparent and interpretable by providing understandable explanations for their decisions and actions \cite{GilpinBYBSK18}. By addressing the opaque nature of many AI algorithms, XAI aims to bridge the gap between the technical intricacies of machine learning and the human need for understanding. Transparency in AI is not only a technical challenge, but also a social and ethical imperative \cite{SabbatiniC24}. Stakeholders need to trust AI systems to interact effectively with them. This is especially critical in sensitive areas where the stakes of incorrect or misunderstood AI decisions are high, such as military applications  \cite{Nitzl2024,NitzlCKB24b}. It is also important to prevent AI from being increasingly integrated into criminal and harmful activities and going undetected \cite{BlauthGZ22}.

The present work contributes to this field by offering a systematic categorization and analysis of symbolic and connectionist approaches within AI. Symbolic approaches are grounded in explicit, rule-based reasoning, while connectionist approaches, exemplified by artificial neural networks, rely on distributed representations and statistical learning. This paper not only delineates the relationships, distinctions, and respective advantages of these approaches but also demonstrates the value of deriving equivalent symbolic decision models from neural models. Such derivations enhance transparency and interpretability without compromising the functionality of the underlying AI systems \cite{ZhangTLT21}.

A special focus is placed on feedforward neural networks \cite{Bebis1994}, including similar types of neural networks that can be represented in a feedforward form, such as convolutional neural networks. Recurrent architectures are not explicitly considered in this study. The analysis underscores that symbolic models such as decision trees, finite state machines (for classification or as transducers), and behavior trees are viable candidates for representing the decision logic of neural networks. Among these, decision trees are identified as the most appropriate symbolic counterpart due to their fit with internal processes, interpretability, and alignment with the explainability goals of XAI \cite{DwivediDNSRPQWS23,CostaP23}.

This paper thus situates itself within the broader discourse on enhancing AI explainability by bridging the gap between connectionist models and symbolic reasoning. By doing so, it contributes not only to the technical development of interpretable AI systems but also to their societal acceptance and ethical deployment. In the following sections, we delve deeper into the methodologies and findings that underpin this contribution, offering insights into the practical and theoretical implications of deriving symbolic representations for neural networks. This work serves as a step forward in the journey toward making AI not just powerful but also accountable and understandable.

The paper is organized as follows.
Sect.~\ref{sec2} introduces basic concepts and reviews key related work in the field, providing both a solid theoretical framework and practical context.
Sect.~\ref{sec3} outlines the core contribution of this work, detailing the methodology for deriving decision trees from feedforward neural networks (FNNs). The process is demonstrated with clear examples and supported by suitable data structures, broken down into simple, easy-to-follow steps. Notably, the final derivation procedure is also applicable to decision paths in deep feedforward neural networks, ensuring scalability and versatility.
Sect.~\ref{sec4} presents a proof-of-concept prototype capable of managing larger FFNs and CNNs with their components, including different layers, pooling mechanisms, and the hierarchical structure of the resulting decision tree. Additionally, a brief overview of the graphical user interface is provided.
Sect.~\ref{sec5} summarizes the procedure for deriving a hierarchical decision tree from a feedforward neural network, highlighting the creation of decision paths for input vectors and their combination into a unified tree, showcasing how symbol-based models make the neural network's inner workings transparent.

\section{Basics and Related Work}\label{sec2}
Artificial intelligence as a generic term for automated, self-optimizing and similar systems should serve as a starting point for the necessary classification of our subject of investigation.
According to \cite{RusselAndNorvig}, the field of artificial intelligence can be divided into the following four categories based on their respective goals:
\begin{itemize}
\item realization of human thought
\item generating human actions
\item realization of rational thought
\item generating rational actions
\end{itemize}
To achieve the last goals, the field of artificial intelligence offers two basic approaches, which are partly contradictory in the way they work. Their competing basic ideas are described by Minsky \cite{Minsky1991LogicalVA}. The first is the \textit{symbolic approach}, which relies on logic-based formalisms that process information with complete, defined symbols. The second is the \textit{connectionist approach}. This is based on the distributed representation and computation of information using many simple mathematical operations. The associated models are usually represented as networks or as a cascade of vector and tensor computations, as in \cite{TFAPI}, among others. The symbol-based approach corresponds to a top-down procedure in which the entire problem is first analyzed, then broken down into manageable, defined subproblems, and then solved. The connectionist approach, on the other hand, is a bottom-up procedure in which the existing model is given predefined information about the problem to be solved and independently generates a sufficiently accurate approximation of the solution over several iterations \cite{Minsky1991LogicalVA}.

The top-down method of the symbolic approach has become established in many technical domains due to the high degree of maturity of the various formal representations of knowledge \cite{AnYHLCL24}. The underlying advantages of this success are, in particular, efficient, systematic search procedures and the reliable management and control of complex configurations of individual elements and complex interactions of subgoals. The disadvantages come into play when the problems to be solved are too unstructured to be described by universally valid axioms, or when these problems are based on analogies and approximations \cite{SeidelSB18}. 

The bottom-up method of the connectionist approach was only able to establish itself on a larger scale when symbolic approaches reached their limits in areas such as pattern recognition, automatic optimization, and clustering, and when large data and computing resources became available. The reason for this, besides the hardware requirements, was the lack of internal structure and architecture of these approaches, which made them unsuitable for classical problems of higher reasoning. The problem of lack of structure has been mitigated somewhat in recent decades by the development of new special forms of artificial neural networks, as can be seen in \cite{Schmidhuber14}. 

\begin{figure}[ht]
\centerline{\includegraphics[width=\columnwidth]{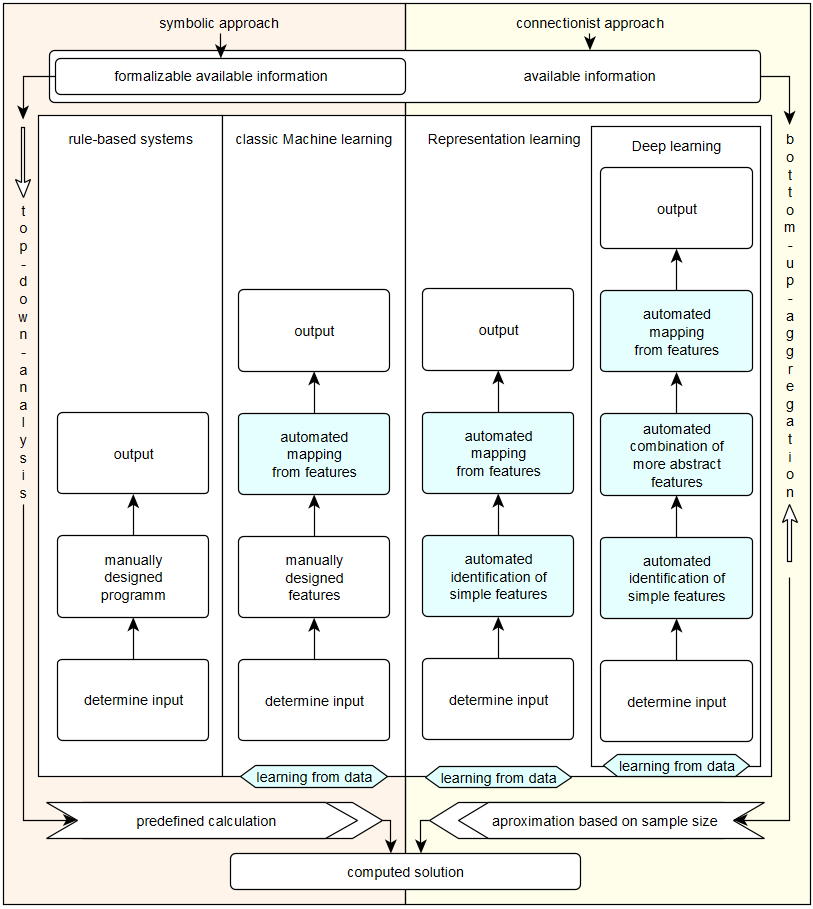}}
\caption{Symbolic vs.\ connectionist approaches.}
\label{seide01}
\end{figure}

Fig.~\ref{seide01} summarizes the division into symbol-based and connectionist approaches according to \cite{Minsky1991LogicalVA} and the categorization of the associated methods and systems according to 
\cite{Goodfellow-et-al-2016}. 

\textit{Rule-based systems} include symbol-based approaches, some of which are thousands of years old, such as classical logic. If a problem is simple and can be formalized well, systems in this category can usually solve it reliably and efficiently. This is done by determining the necessary input data, manually constructing an effective procedure for solving the problem, and then determining the desired solution outputs. Examples of methods in this category include inference machines \cite{BrowneS01} and reasoning in first-order logic \cite{BaderHH08}.

\textit{Classic machine learning} is capable of making assignments on its own based on given, defined characteristics. This category also includes symbol-based approaches. A recent example of this category is decision tree learning, as described in \cite{geron2022hands}.

It was not until the early 2000s that \textit{learning from representations} could be transferred from theory to an efficiently usable practical application, as described in \cite{Schmidhuber14}. Initially, the input data is determined manually, but then the characteristics of the problem to be solved are determined independently by the system used. Subsequently, the data is independently reorganized and mapped by the respective system on the basis of the determined characteristics in the recognized problem context, and finally the desired solution is displayed. Perhaps the best known example of this category of AI systems are flat artificial neural networks, which are based on the idea of the \textit{perceptron} first described in \cite{Rosenblatt}.
As Goodfellow \textit{et al.} \cite{Goodfellow-et-al-2016} explain, connectionist approaches were called cybernetics from 1940 to 1960, and much later deep learning, where the aspect of \textit{depth} of the artificial neural model is not exactly quantified. They refer to multi-layer artificial neural networks as deep neural networks; the term artificial neural networks can be considered as an alternative term for methods and models that learn representations based on the connectionist approach.

\textit{Deep learning} evolved in the late 2000s as the availability of data, storage, and especially computing power continued to increase. This computing power was generated by new hardware, especially graphics cards optimized for vector computations. By providing it in large data centers organized according to the cloud principle, computing power became available more or less independently of location. In addition, motivated by the successes of the early 2000s, a large number of specialized connectionist models were created or old specialized models were made usable in practice. These developments are described in \cite{Schmidhuber14}. 
The defining feature of deep connectionist models in deep learning is the high number of successive, distributed processing steps. This makes it possible to solve much more complex problems with multiple levels of correlation. However, the aggregation of problem contexts and decision paths is usually beyond human comprehension. 

At this point at the latest, questions are being asked about the trustworthiness and explainability of AI \cite{Hagras18,RawalMRSA22}.
Phillips \textit{et al.} \cite{XAI2Nist2021} specify four principles of XAI, see also \cite{AIExecutiveOrder} and \cite{AIRMF}, where the explainability and interpretability of AI models is listed as a relevant property of artificial intelligence. 
The goal of explainable AI is described among others in \cite{XAIR} and \cite{XAIDarpa}. Both sources also make clear that the successful use of models from the field of machine learning, and especially those connectionist models from the field of deep learning, makes it necessary to explicitly consider the explainability of AI models. 

We will also use the term \textit{artificial neural network} (ANN) for the connectionist approach according to \cite{Minsky1991LogicalVA}. Both terms describe the same area of research. 
ANNs have the ability to learn internal representations for these problems by independently determining and assigning problem characteristics. This usually results in probabilities or probability distributions as a statement about a calculated decision. The internal representations are learned by the corresponding artificial neural networks, which optimize the determination and assignment of features by adjusting their connection weights. To do this, they propagate the deviation of a decision result from the desired result back proportionally through their structure and correct the error portion that each weight had \cite{ParallelDistributedProcessing}. The numerical method used for this iterative optimization is usually the gradient descent.
For this self-optimization in the event of wrong decisions, there is \cite{Dike2018}
\begin{itemize}
\item \textit{supervised learning}, where a suitable output value must be marked or labeled for each individual input when training the network. The provision of data is more complex, but the self-optimization is more reliable and converges faster to a local optimum;
\item \textit{reinforcement learning}, where at the end of an individual decision, the only thing that is checked is whether it was useful according to a reward function or not. The weights are then either strengthened or weakened depending on their contribution to the decision. Sufficient data is generated more quickly, but the optimization process is generally less reliable and requires more iterations than supervised learning.
\item \textit{unsupervised learning}, where the goal is not to optimize the correctness of a decision, but to optimize the representation for all inputs. A common example is the assignment of inputs with $n$ features according to their spatial proximity with respect to these features, e.g.\ using $k$-means. The data is grouped according to a predefined similarity.
\end{itemize}

\begin{figure}[ht]
\centerline{\includegraphics[width=\columnwidth]{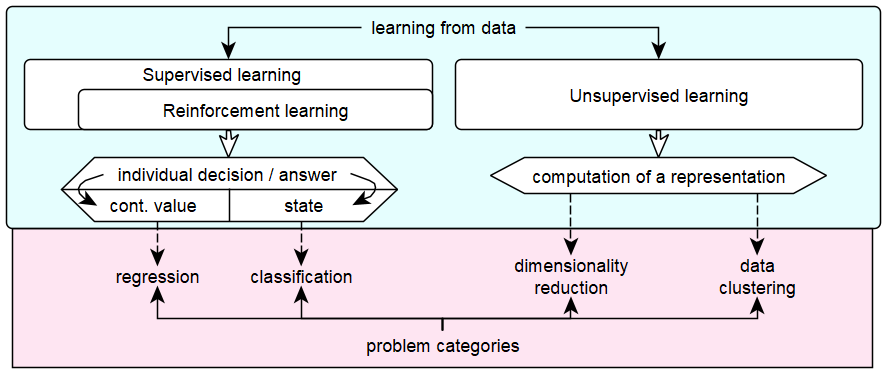}}
\caption{ANNs' learning approaches and the problem categories in focus according to \cite{geron2022hands}.}
\label{seide02}
\end{figure}

Fig.~\ref{seide02} compares these three learning approaches and assigns them to the categories of problems that are handled by the artificial neural networks that typically use each learning algorithm. 

\begin{figure}[ht]
\centerline{\includegraphics[width=\columnwidth]{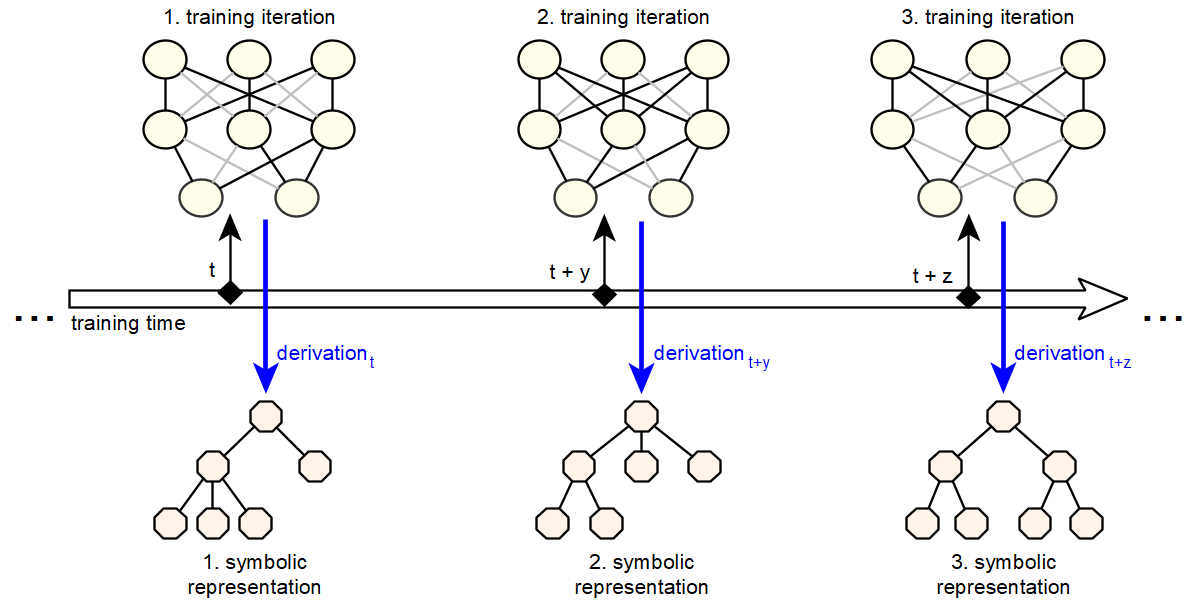}}
\caption{Training-time dependent derivation of equivalent symbol-based decision representations.}
\label{seide03}
\end{figure}

As shown in Fig.~\ref{seide03}, learning internal representations by iteratively optimizing the connection weights using the gradient descent method generates different training iterations of an artificial neural network, each with its own iteration of the learned internal representation. In this way, the symbol-based model representation of a network at time $t$ can be different from that at later times $t+y$ and $t+z$, where additional training intervals and thus weight adjustments have been performed. 

\begin{figure*}[t]
\centerline{\includegraphics[width=.65\textwidth]{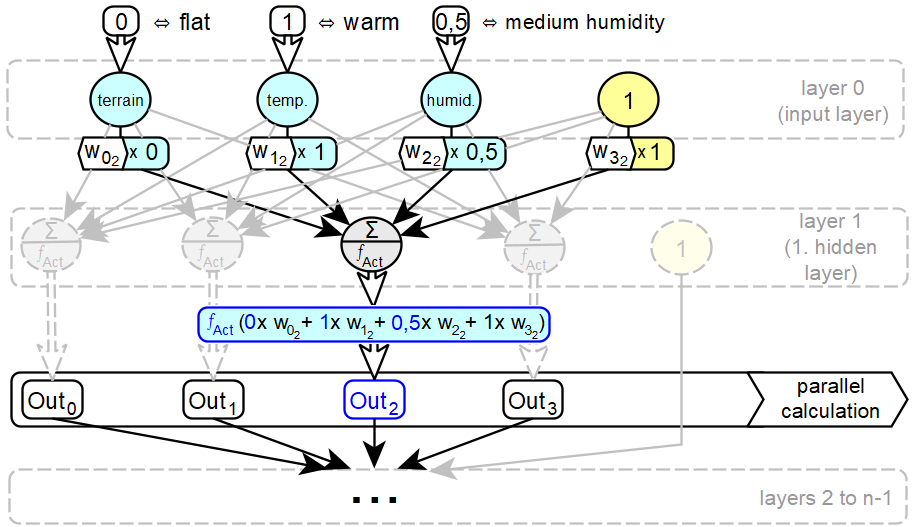}}
\caption{Basic principles of parallel information processing in ANNs with hidden cells.}
\label{seide04}
\end{figure*}

Learned representations of ANNs can be generalized based on similarities \cite{ParallelDistributedProcessing}. ANNs are divided into groups of neurons, called layers, which are interconnected. This means that each layer can be viewed as a function of its predecessor layer. Thus, there is a coherent distributed representation of information within each layer. From one layer to its successor layer, the entire distributed representation is modified using a transfer function and, in conjunction with the internal functionalities of the successor layer, forms a new coherent distributed representation. The layered distributed representation of information is reflected in the fact that the locally represented input values in the input neurons of a network in each subsequent layer have a proportional contribution to the activation of the contained neurons of that layer, thus representing their distributed activation contributions. This property is exploited in modern frameworks, such as \textsc{TensorFlow} \cite{TFAPI}.

Fig.~\ref{seide04} shows an ANN with $n$ layers and the input parameters for altitude, temperature, and humidity. In addition to temperature, input information is given for a flat altitude, coded $0$, and for a medium humidity, coded $0.5$. In layer 1, the input values of the neurons $i$ of layer $0$ multiplied by the corresponding connection weight $w_{i_j}$ are processed in parallel by all neurons $j$ of this layer as part of the input activation of the neuron $j$. The input activation in the second neuron of layer 1 consists of a $w_{0_2}$ component of the representation $0$ for the information flat, a $w_{1_2}$ component of the representation $1$ for the information warm and a $w_{2_2}$ component of the representation $0.5$ for the information wet. In this step, the connection between the two aspects of artificial neural networks---distributed representation of information and parallel processing of information---becomes clear. Next, this input activation is converted into the output activation of the respective neuron in the layer, in this case the second neuron, by the activation function $f_{Act}$. These two steps for calculating the output activation $Out_j$ of each neuron $j$ of layer 1 are performed in parallel in each neuron $j$ in this order. A complete processing cycle through a network for a complete set of inputs is called a \textit{forward pass}. 

\textit{Convolutional neural networks} (CNNs) are specialized for image recognition, speech recognition and language processing. The basic idea behind this type of network comes from the realization in \cite{Hubel1962ReceptiveFB} that the neurons in the visual cortex of animals have only a local field of perception and thus only react to stimuli from a small area of the entire visual field. This later led to the development of a neural model with the \textit{neocognitron} described in \cite{Fukushima1983NeocognitronAN}, which pursued the idea of processing only the activations of spatially adjacent neurons by their successor neurons and thus aggregating individual image sections layer by layer into increasingly extensive representations of partial images. 
CNNs and the associated properties of their elements can be considered as a special form of feedforward neural networks and can be fully represented by hidden cells, taking into account the relevant specifics \cite{BachB2015}.

In the following, the derivation of \textit{decision trees} as equivalent symbol-based decision models for given feedforward (or convolutional) neural networks will be developed step by step. 
Other generative artificial neural networks such as Hopfield networks, Boltzmann machines, autoencoders, and long / short term memory networks and their specific elements \cite{Veen2016} are beyond the scope of the derivation to be developed in this work. 
The research methodology underlying this work is a thorough and systematic investigation of symbol-based and connectionist AI approaches. By providing a prototype implementation, we can demonstrate practical feasibility. 

\section{The Derivation of Decision Trees}\label{sec3}
Before we start with the derivation, a few terms need to be defined in more detail.

\subsection{Symbols and combination rules}
A \textit{symbol} is a single entity with a defined meaning. Finite sets of these entities can be combined according to defined rules to form new entities, to which a new, defined meaning is then assigned. Thus, there are the following categories of symbols: complete entities, which have a meaning assigned by definition, and composite entities, whose meaning results from their components and their meanings on the one hand, and from the rule for combining these components on the other.
Whether an entity is complete or composite depends on the permitted \textit{combination rules}.

When determining the category of a symbol, the permissible set of combination rules must always be taken into account. For example, a letter is a complete symbol if only the combination rules of the respective written language are to be considered. However, if combinations in the sense of pictorial representation are to be included, a letter can be considered a composite symbol consisting of a set of points.

\subsection{Fillers and roles in feedforward networks}
Fillers are assignments, i.e. perceptible representations, for an associated role in a symbol-based overall structure. They are thus a representation of a symbol entity, be it a complete or a composite symbol.
Similarly, roles can be considered as the relations or relationships of a symbol entity represented by a filler to all other symbol entities present in a common symbol-based structure.
The combinations of fillers and roles are realized in symbol-based information processing by the respective valid combination rules. In this context, the combination rules always refer to the roles, since the fillers are only one, basically freely selectable representation.

This idea was originally presented in \cite{smolensky2006harmonic}, where symbols consisting of fillers and roles are depicted by numerical activation values forming specific activation patterns.
Their combination is then realized as a superpositioning of these vectors in an intermediate vectorial level. The result of this operation can again be regarded as a complete unit, comparable to a single symbol or a symbol-based structure $s$, where 
$s = \sum_{i}^{} f_i \otimes r_i$, the sum total of all combinations of fillers $f_i$ and their associated roles $r_i$ contained in $s$.
The (\emph{filler}) $f_i$ represents a clearly assignable sign.
The structural role $r_i$ describes the relationships of the sign to all other signs in the overall symbol $s$.

\begin{figure}[ht]
\centerline{\includegraphics[width=\columnwidth]{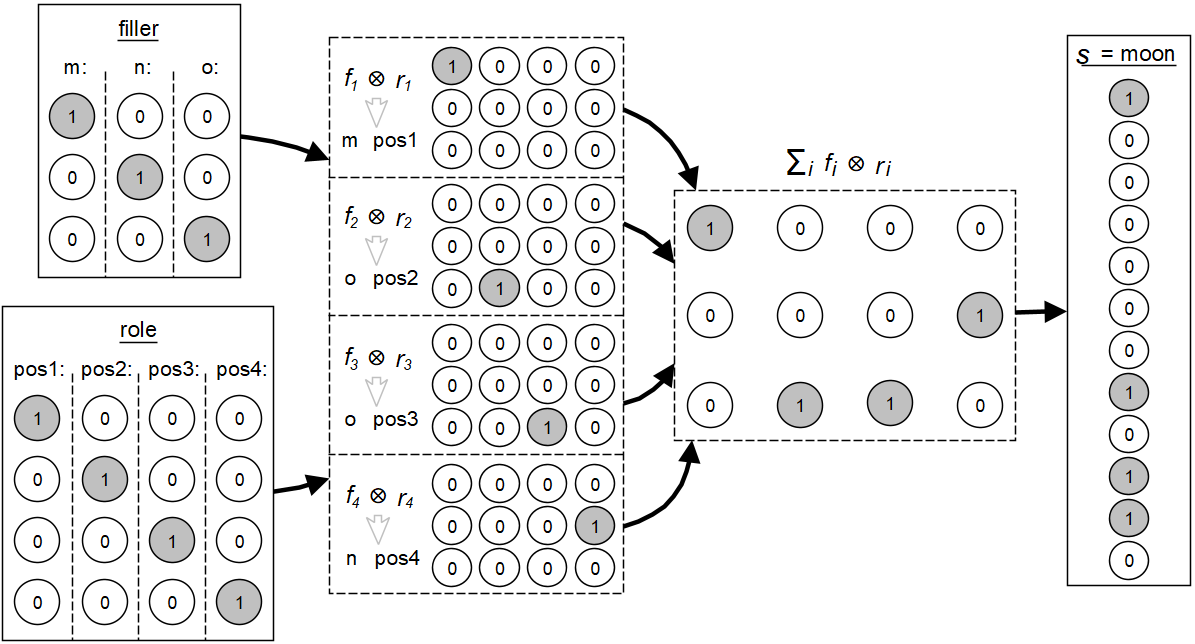}}
\caption{Example of the concept of fillers and the roles in the integrated connectionist / symbolic architecture according to the explanations in \cite{smolensky2006harmonic}.}
\label{seide05}
\end{figure}

\begin{figure}[ht]
\centerline{\includegraphics[width=\columnwidth]{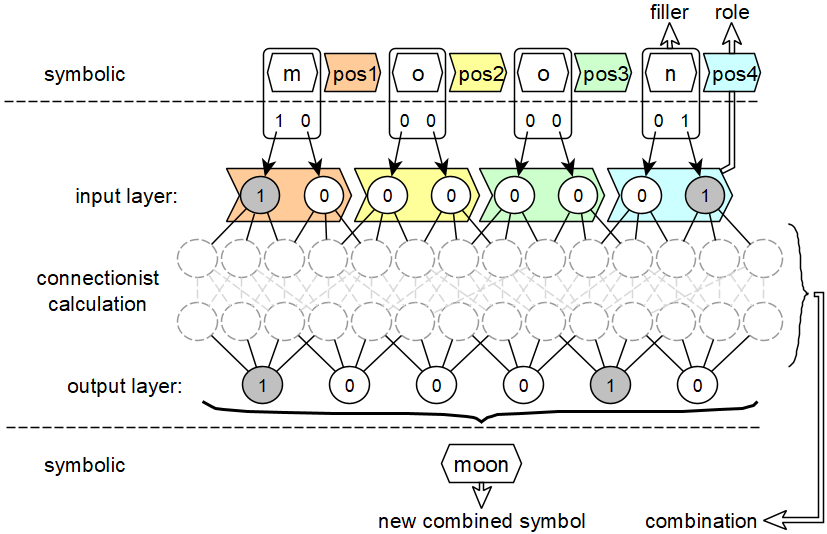}}
\caption{Intended use of the concepts of fillers, roles and combinations of activation patterns from the integrated connectionist / symbolic architecture applied to feedforward neural networks to derive symbol equivalents.}
\label{seide06}
\end{figure}

A simple example is any word $s$ with its specific meaning, formed from all contained letters $f_i$ and their respective positions $r_i$ in that word. 
Fig.~\ref{seide05} illustrates this using the symbol structure associated with the  word \textit{moon} according to the concept of the integrated connectionist / symbolic architecture from \cite{smolensky2006harmonic}. 
The fillers, here the letters \textit{m}, \textit{n} and \textit{o}, are each formed by 3-dimensional vectors consisting of binary values. 
The roles, here the positions at \textit{pos 1}, \textit{pos 2}, \textit{pos 3} and \textit{pos 4} in the word, correspond to 4-dimensional vectors of binary values. The summed tensor products of the vectors result in a 3$\times$4-dimensional matrix $s$, which in turn can be represented as a 12-dimensional vector. 
This matrix and this vector are again two representations of the word \textit{moon}.
This example also illustrates the connection between the symbol-based, vectorial, and neural levels mentioned above. 
The transfer of this principle to feedforward neural networks is shown in Fig.~\ref{seide06}. 

The symbolic level, which is adjacent to the input layer in the form of the interpreted input, contains compressed carriers of information meaning, consisting of their filling word, the letter, and their role, the position in the word. 
The neural layer, realized by the neural network, describes the connectionist derivation of output activations from input activations using activation patterns that implement this derivation. 
The respective output activations and their associated dependencies on the activation values of the inputs describe how the new overall symbol is formed from the fillers and roles of the individual symbols of the input. 

While fillers and roles for the generation of symbols can be determined from the inputs of a feedforward neural network, if necessary by inverting the preprocessing of these inputs, the combination of these fillers and roles is found in the structure of the respective networks. A single combination is realized by the incoming weighted connections $w_j$ and the corresponding transmitted activation values $v_j$ of a single neuron in a subsequent layer.

This layer follows each layer in the examined feedforward network for whose neurons valid symbol equivalents with associated fillers and roles have already been determined. The input value for each neuron is determined by the sum $\sum_{i} v_i \times w_i$. 
This sum represents the function, also called pattern or rule, 
by which the composite symbol associated with the neuron in question is determined in relation to its predecessor neurons. 
All symbols with their associated fillers and roles are potentially considered for those combinations that are associated with a predecessor neuron that can be reached via an incoming connection $j$. If the product $v_j \times w_j$ associated with $j$ is sufficiently relevant for the activation $f_\text{Act}(\sum_{i} v_i \times w_i)$ with $j\in \{i\}$, then the symbol of the neuron for the incoming connection $j$ is relevant for the combination and will be considered for it. 
After the combination, the resulting symbol for the target neuron consists of the set of all tuples $(\text{filler}_j, \text{role}_j)$ whose associated product $v_j \times w_j$ is sufficiently relevant for $f_\text{Act}(\sum_{i} v_i \times w_i)$ with $j\in \{i\}$. The output activation $f_\text{Act}(\sum_{i} v_i \times w_i)$ calculated in the target neuron must be sufficiently relevant for the input activations of the successor neurons, so that the resulting symbol is considered for the combination with the linked symbol of the respective successor neuron.

Since it is possible with the steps described above to generate both fillers and roles, as well as their combinations, from individual elements of the complete computation of a decision in a feedforward network with associated inputs and outputs, these concepts can be used in the further course of this work as bridging elements between feedforward networks and decision models that use symbols. 
The idea here is to first determine the fillers and roles for the calculation of a decision and then the respective valid linked sets of tuples $T = (\text{filler, role})$ and thus to generate the necessary symbol equivalents for a symbol-based decision model equivalent to the examined feedforward network.

\subsection{Decision trees and feedforward neural networks}
To derive an equivalent decision tree from a given neural network, the structure of the underlying graph must be created in an additional step. This graph connects the later symbol equivalents. However, there are several slightly different representations of decision trees in \cite{geron2022hands}, \cite{Millington-Funge-2009}, \cite{DecTreeDef}, and \cite{SeidelB18}.

A decision tree is a directed graph. It starts in exactly one node, ends in decisions $D$ and consists of nodes $N$ and edges $E$.
The nodes $n \in N$ are described as tuples $n = (s, t)$ where
$s$ is a defined information state $s = \{i_0, \ldots, i_{k-1}\}$ consisting of $k$ individual items of information $i_j$ with $j \in \{0, \ldots, k-1\}$. Each individual item of information $i_j$ is assigned a set of $m$ possible values $b_{i_{j_r}} \in \{i_{j_0}, \ldots, i_{j_m-1}\}$ with $r \in \{0, \ldots, m-1\}$ assigned to each piece of information $i_j$. The assignments are represented by symbols.
$t$ is a test on at least one piece of individual information $i_j$, so that $s \times t \times b_{i_{j_r}} \rightarrow e_r$ applies with edge $e_r \in E = \{e_0, \ldots, e_{m-1}\}$ and $m =$ number of initial edges.
The edges $e$ are described by their connection destination $g$. Here, $g \in (N \cup D)$. Each edge is linked to the validity / occurrence of an assigned allocation.
The decisions $d \in D$ are described by their specific output value.

\begin{figure}[htb]
\centerline{\includegraphics[width=\columnwidth]{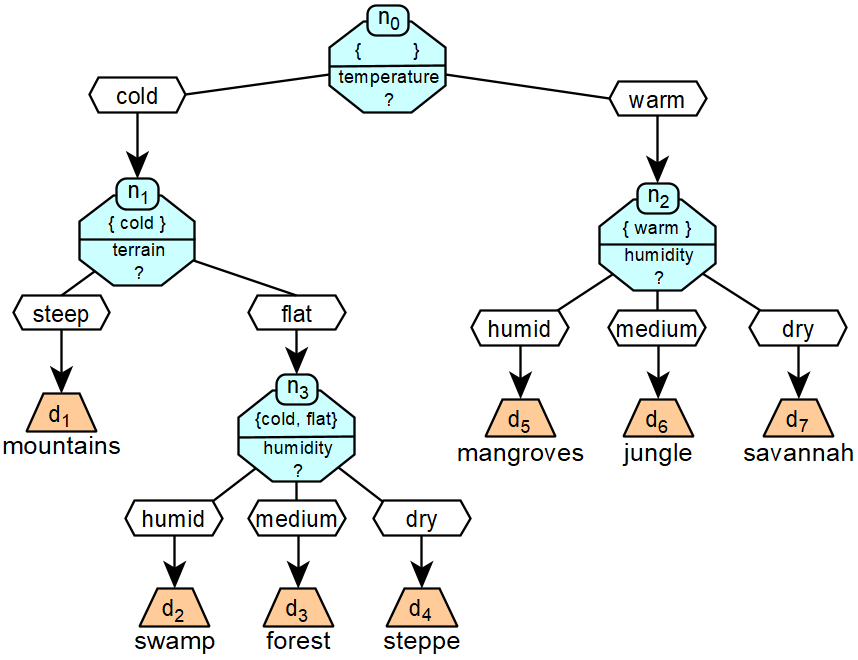}}
\caption{A concrete realization of a decision tree that corresponds to the formalized example introduced in Fig.~\ref{seide04}.}
\label{seide07}
\end{figure}

Fig.~\ref{seide07} shows an example of a specific decision tree. 
The decision tree classifies a given landscape into the seven categories mountain, swamp, forest, steppe, mangrove, jungle and savannah, which are described by the respective decisions $d$. The classification is based on information $i_j$ about the qualitative temperature ($i_0$), altitude ($i_1$) and humidity ($i_2$) of the landscape. The possible values of $b_{i_0}$ are cool and warm, $b_{i_1}$ flat and steep and $b_{i_2}$ wet, medium and dry.

\begin{figure}[ht]
\centerline{\includegraphics[width=\columnwidth]{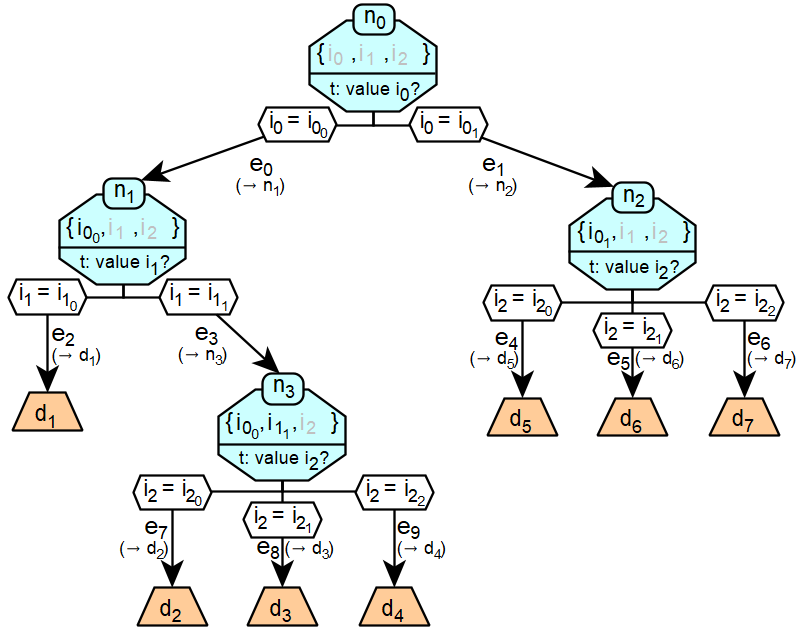}}
\caption{Example of a formalized decision tree.}
\label{seide08}
\end{figure}

Fig.~\ref{seide08} shows a more formalized version of the decision tree from Fig.~\ref{seide07}.

\begin{figure*}[t]
\centerline{\includegraphics[width=.8\textwidth]{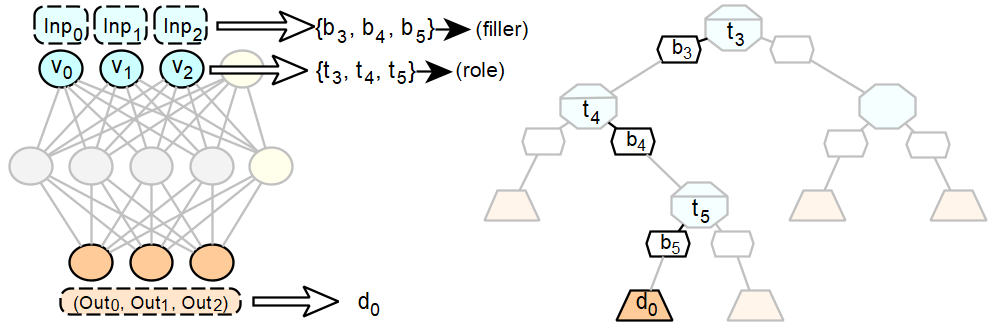}}
\caption{Exemplary representation of the elements of a single path of a decision tree that can be derived from a single run of a given feedforward neural network.}
\label{seide09}
\end{figure*}

\subsection{Identification of symbols and their relationships to define the structures to be derived}
In the nodes $N$ of a decision tree, a test $t$ contained in the node is assigned a certain possible assignment $b_{i_{j_r}}$ to a single piece of information $i_j$ belonging to that test, or to a certain set of such single pieces of information. The assigned mappings $b_{i_{j_r}}$ follow the principle of fillers associated with their respective roles. So they are assignments with symbols. The information states $\{i_0, ..., i_{k-1}\}$, over which the tests $t$ are performed, correspond to the principle of possible roles to be assumed. Together with an associated filler from an assignment, they form a complete symbol. A defined connecting edge $e_r$ is always connected to this symbol $\{i_0, ..., i_{k-1}\} \times b_{i_{j_r}}$ and thus the following node $n$ or the corresponding decision $d$ of this edge. The symbols $\{i_0, ..., i_{k-1}\} \times b_{i_{j_r}}$ can be found in decision trees in the edges $e$ when they appear for the first time in response to a test $t$, as well as in the information states $s$ of all nodes $n$ following the said $e$. 

The structure of a decision tree can thus be represented by the symbol components mentioned. 
Deriving these components from a neural network is a prerequisite for generating an equivalent decision tree for a set of decisions computed by that neural network. 
These components can be used to generate both the structure of the tree and the required symbolism.

Therefore, the components needed to determine how to structure a decision tree are as follows.

\begin{itemize}
\item the test $t$ checked in the respective node over a defined set of individual pieces of information and the associated known information states $\{i_0, \ldots, i_{k-1}\}$. The information states correspond to already existing/valid combinations of fillers and roles. The tests are used to determine valid fillers for roles that have not yet been filled.
\item the possible combinations of assignments $b_{i_{j_r}}$ after the respective node, which represent a valid answer to the test of the node. These correspond to the existing/valid combinations of fillers and roles after the test.
\item the edges $e_r$ associated with the valid combinations of assignments to the respective subsequent nodes or subsequent decisions.
 \end{itemize}

In order to determine these components, we will now take another look at the elements offered by the basic form of neural networks.

The edges $e_{i_j}$ run in a directed manner between an input neuron $i$ and a target neuron $j$ and each has a weight $w_{i_j}$. This weight is a numeric value that is multiplied by all activation values $v_i$ of the input neuron $i$ of the edge that are passed along this edge.
Input neurons a have an input value $Inp$ and a set of $n$ output edges $\{e_{a_0}, \ldots, e_{a_{n-1}}\}$.
Hidden neurons b have a set of $m$ input edges $\{e_{0_b}, \ldots, e_{(m-1)_b}\}$ and a set of $n$ output edges $\{e_{b_0}, \ldots, e_{b_{n-1}}\}$. Furthermore, each of these neurons has an activation function $f_\mathrm{Act}$ and an input function, which usually corresponds to the sum function.
Output neurons c have a set of $m$ input edges $\{e_{0_c}, \ldots,$ $e_{(m-1)_c}\}$ and an activation function $f_\mathrm{Act}$ as well as an input function, which usually corresponds to the sum function. In addition, each output neuron has an output value $Out$.

When a neural network is created, it is determined which input neurons encode a particular piece of information. The coded information is therefore always known, as can be seen from \cite{Goodfellow-et-al-2016} and \cite{geron2022hands}, among others. This coded information corresponds to the individual information $i_j$ from decision trees. This also allows to derive the tests $t$, i.e. the information $i_j$ on which these tests have to be performed. The assignment of the tests to a node and the associated information states is not yet possible. The respective combinations of input values $Inp$ correspond to the specific assignments $b_{i_{j_r}}$ of the information $i_j$ of a decision tree, which apply to a defined, associated decision $d$. These combinations thus also realize the concept of fillers according to \cite{smolensky2006harmonic}. The coding of the possible decisions of a neural network by the output values $Out$ of the output neurons is also known, since this is also determined when the neural network is created. 

Each output vector of a neural network is associated with a specific decision, which is equivalent to the decisions $e$ of the decision trees. The individual inputs for the coded information, which correspond to specific assignments $b_{i_{j_r}}$, and the resulting output of the neural network, which corresponds to a single decision $d$, are each assigned to a decision process of a neural network that has been run through. Their equivalents can therefore be regarded as individual elements of a specific path through a decision tree, hereafter referred to as a decision path, which leads to exactly the decision $d$ that corresponds to the output of the neural network for the decision made. 

See Fig.~\ref{seide09} for an example.

In order to be able to make statements about the arrangement of the pairs from a test $t$ and the associated subsequent assignment $b_{i_{j_r}}$, it is first clarified which properties are associated with the position of such a pair at a certain point in a decision path. Then, the internal structure of the neural network used for the decision is taken into account in order to determine subgraphs that allow conclusions to be drawn about the desired properties. In doing so, the peculiarities of the processing of individual pieces of information $i_j$ and their values $b_{i_{j_r}}$ to generate a decision in neural networks must be taken into account. In particular, the distributed representation of information must be considered. The properties to be considered for the position of a pair $(t, b_{i_{j_r}})$ are: 

\begin{itemize}
\item once a pair $(t, b_{i_{j_r}})$ has an assigned position in a decision tree, the assignment $b_{i_{j_r}}$ must be the valid assignment for the associated information $i_j$, which applies to all subsequent decisions $d$ after this position. In particular, this follows from the approach $s \times t \times b_{i_{j_r}} \rightarrow e_r$ for determining the successor edges, since the decisions $d$ are at the end of an associated cascade of edges.
\item there is a direct dependency of the decisions $d$ on the preceding valid assignments $b_{i_{j_r}}$.
\item there is a direct dependency among the assignments $b_{i_{j_r}}$ of a single piece of information $i_j$. Once a position in a decision tree has been determined for an assignment of a piece of information, no other possible assignment for the same information can follow after this position in the respective decision path.
\end{itemize}

The assignments of the input values $Inp$ are not recorded and processed in a single place in the hidden layers that follow the input layer, but are distributed in different places. This means that they are distributed and proportionally included in the determination of the overall decision. In principle, each neuron corresponds to a function that converts an input vector into a scalar. The input vector in turn contains the scalars of the previous layer as components, which for the neurons of the first hidden layer are the concrete numerical values of the respective $Inp$ values. Thus, with increasing depth, the hidden layers contain more and more partial representations or assignments of an $Inp$ value until the final decision is made. 

The output $Out$ of a neuron is 
$Out = f_{Act}(\sum_{i} v_i \times w_i)$ with $v_i$ = input value $i$ and $w_i$ = weighting factor for input value $i$.
The value $Out$, which is either an intermediate or a final result of the decision calculation, always depends on the sum of the total input and thus on the complete input vector over all weighted input values (proportional occupancy). The basis for this is again the distributed representation of information in artificial neural networks combined with parallel processing of this information. In multilayer neural networks, this leads to concatenated or cascaded combinations of the original input values across the successive layers of such a network. For a neural network with an output layer $n$, this corresponds to the form $Out = f_{Act_n}(\sum_{i} (f_{Act_{n-1}}(\sum_{j} (...) \times w_j)) \times w_i)$.

\begin{figure}[ht]
\centerline{\includegraphics[width=\columnwidth]{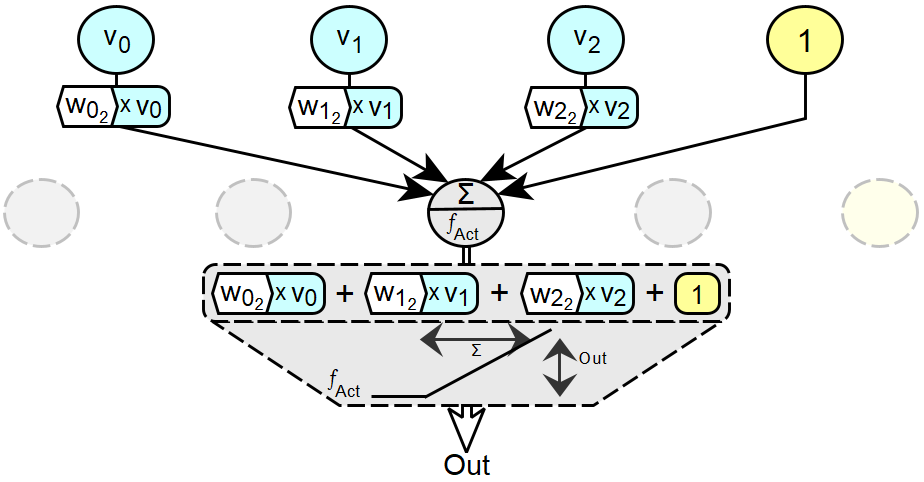}}
\caption{Exemplary visualization of the principle of dependence of individual occupancy values on valid partial occupancies at specific locations (neurons) in neural networks.}
\label{seide10}
\end{figure}

The validity of a partial input value at a specific location in a neural network as part of the calculation of a desired output always depends on the simultaneous validity of all other partial input values at that specific location which, in combination with the input value under consideration, lead to the desired output. This is illustrated in Fig.~\ref{seide10}.

The properties of feedforward neural networks just described are essential for the process of input processing to compute a decision and to determine the symbol components fillers, roles and their combination in the structure of feedforward networks and assign them to the respective neurons. First, we have to determine which proportional distribution of the assignments of the input values in a single processing step contributes to the calculation of the output vector in the output layer and thus to the generation of the resulting decision. The relevant proportions of the distributed representation are passed from the penultimate layer to the output layer. Therefore, the neurons of the penultimate layer of the considered network that send a value of sufficient magnitude to the output layer to make a relevant contribution to the generation of the overall output should be examined.
For each of these neurons $j$ of the penultimate layer, it must be determined for each input neuron whether this value makes a sufficiently large numerical contribution to the input sum $\sum_{i} v_{i_j} \times w_{i_j}$ of neuron $j$ in order to be relevant for the activation of the neuron. 

The following steps are carried out to capture relevant distributed representations of input configurations in a decision path for the computed decision $d$.

\begin{enumerate}
\item if a neuron $j$ has no connection to a neuron $k$ in the output layer through which a sufficiently large weighted value $v_{j_k} \times w_{j_k}$ that is significant for the calculated output activation, then the neuron $j$ is not considered for the decision path in a decision tree.
\item if the input assignment of an input neuron is not sufficiently relevant for the output activation of a neuron $j$, because there are no connections with sufficient weight to the neuron $j$, then the input assignment of this input neuron is not considered for the neuron $j$ in the considered decision path in a decision tree.
\item otherwise, the corresponding input assignment for the respective neuron $j$ is taken into account in the considered path of the decision tree.
\end{enumerate}

\begin{figure*}[ht]
\centerline{\includegraphics[width=.8\textwidth]{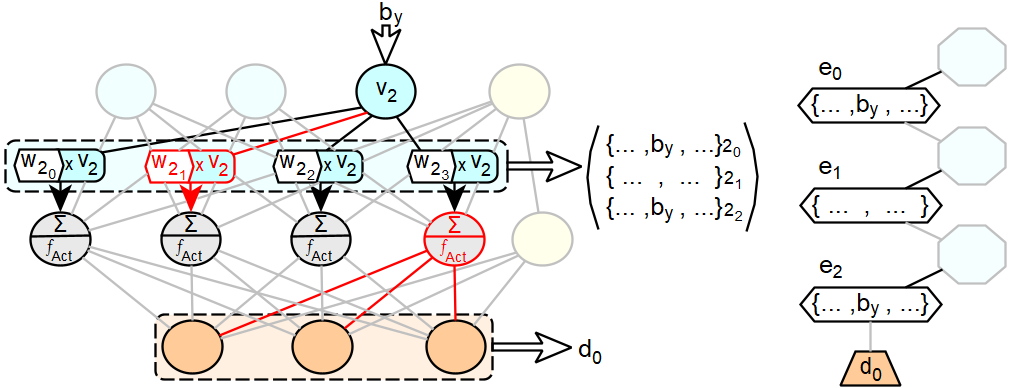}}
\caption{Exemplary illustration of the concept of integrating distributed representation and processing of feedforward neural networks into extended decision trees.}
\label{seide11}
\end{figure*}

The arrangement of the neurons of the penultimate layer is arbitrary, since they are simultaneously considered as neurons of the same layer in the processing of the feedforward neural network. A visualized example of the described procedure is shown in Fig.~\ref{seide11}. It shows a small feedforward neural network with an input layer consisting of three neurons and a bias neuron, a hidden layer with four neurons and a bias neuron, and an output layer consisting of three neurons. The last neuron of the penultimate layer has no connections through which a relevant activation value for the current input to the network is passed to a neuron of the output layer, resulting in the output $d_0$. Consequently, the set of input configurations that produce an activation in this neuron is not considered in the edges of the decision path for $d_0$. The other three neurons of the penultimate layer are considered. However, the second neuron of the penultimate layer has no connection through which it can receive an activation value that is relevant for the current input and that includes the value $b_v$ of the third input neuron. Therefore, $b_y$ is not considered as an allocation for the edge derived from the second neuron.

\subsection{Extension of decision paths for use in deep feedforward neural networks}
Modern feedforward neural networks typically have multiple hidden layers. These layers, like the penultimate layer of a feedforward neural network, represent parts of the processing of the network's input assignments into the corresponding network output. Therefore, to fully account for the aforementioned processing by a derived decision path, all hidden layers of the considered network must be considered in that path.

These processing steps are represented by the subnetworks of the considered network, which start with all input neurons and end with the individual neurons assigned to the edges $e$. From these subnetworks decision paths can be derived, which replace exactly the edge $e$ in the decision path already derived from the whole network, which was derived from the target neuron of the considered subnetwork. This procedure can now be repeated layer by layer, and in these layers neuron by neuron, until each hidden neuron of the analyzed feedforward neural network is represented by at least one associated edge in the derived decision path.

\begin{figure*}[ht]
\centerline{\includegraphics[width=.7\textwidth]{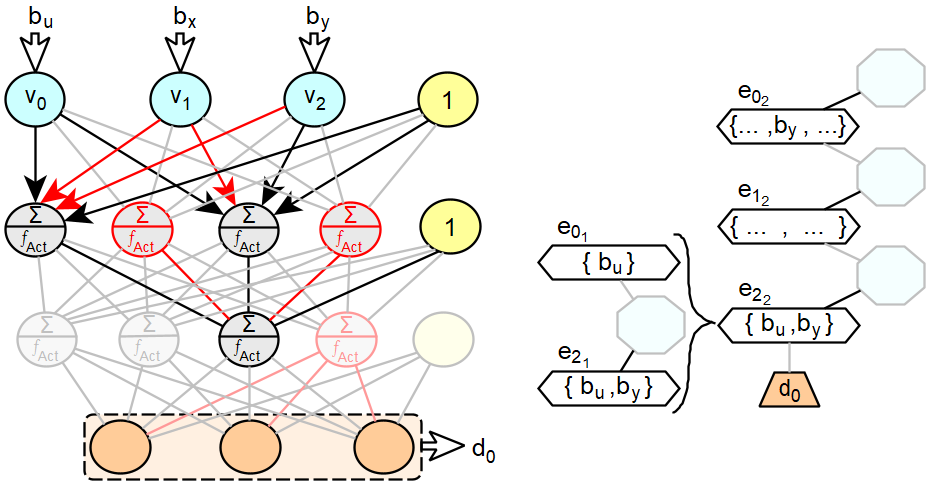}}
\caption{Exemplary illustration of the expansion of decision paths in extended decision trees for feedforward neural networks with several intermediate layers.}
\label{seide12}
\end{figure*}

Using Fig.~\ref{seide12} as an example, we will describe the procedure for a feedforward neural network with two hidden layers. For the third neuron of this layer, the relevant processing steps of the second layer are to be taken into account. In the example shown, the said third neuron of the second layer receives the sufficiently large weighted activations $v_0 \times w_{0_2}$ from the first neuron and $v_2 \times w_{2_2}$ from the third neuron of the second layer. 

Accordingly, the new edges $e_{0_1}$ for the first neuron and $e_{2_1}$ for the third neuron are created in the new subpath associated with the edge $e_{2_2}$. The absolute values of the weighted activations of the second and fourth neurons of the second layer are too small and thus insignificant for the overall input activation of the third neuron of layer three. For the neuron assigned to the edge $e_{0_1}$, only the weighted activation of the input neuron $v_0$ is sufficiently large in absolute value to be relevant for its overall input activation. Therefore, only the input weight $b_u$ of $v_0$ is assigned to the edge $e_{0_1}$. For the neuron associated with the edge $e_{2_1}$, the weighted activations of $v_0$ and $v_2$ are relevant due to their magnitude. Therefore, the input weight $b_u$ of $b_y$ is assigned to $e_{2_1}$. In the following, a procedure for deriving decision paths from feedforward neural networks with more than one hidden layer is described, based on the statements made so far.

We now have all the ingredients for the final derivation procedure.
At the beginning, the output mapping and the associated decision are computed for the given vector of input mappings. At the end of this calculation, the respective output activations $v_j$ generated by the given input vector are also computed for all neurons $j$ of the hidden layers and the output layer. 

Then, for a feedforward neural network with $1$ to $n$ hidden layers, the following steps are to be carried out for each hidden layer $s$ with $s \in \{1, \ldots, n\}$ in descending order starting from layer $n$ for all $m$ neurons of the respective layer $s$. 
Here, the neuron $j$ with $j \in \{0, \ldots, m-1\}$ is the neuron of layer $s$ currently being considered:

\begin{enumerate}
\item if the successor layer of the neuron $j$ is the output layer, then it is necessary to check if there is a connection to a neuron $k$ of the successor layer through which a sufficiently large weighted value $v_{j_s} \times w_{j_k}$, 
which is significant for the calculated output activation, is propagated. If this is the case, the neuron $j$ of layer $s$ is considered with a new associated edge $e_{j_s}$ at the end of the decision path to be derived.
\item if the successor layer of the neuron $j$ is a hidden layer, it is necessary to check whether there is a connection to a neuron $k$ of the successor layer, which is already considered with an edge $e_{k_{s+1}}$ in the decision path to be derived, and whether a sufficiently large weighted value $v_{j_s} \times w_{j_k}$ is transmitted over this connection, which is significant for the calculated activation of $k$. If this is the case, the neuron $j$ of layer $s$ is considered with a new associated edge $e_{j_s}$ at the end of the partial path to be derived, which is assigned to the already existing edge $e_{k_{s+1}}$.
\end{enumerate}

Once the described procedure has been completed, a complete derived decision path with all considered decision levels, i.e. layers of the original feedforward neural network, is available for the considered decision. A single edge $e$ of this path is either directly assigned a set of relevant input configurations or a subpath contained in it. If $e$ is assigned a subpath, the associated set of relevant input configurations is formed from the union of all sets of relevant input configurations of the edges of the assigned subpath.

\section{The Prototype}\label{sec4}
The prototype was created using JDK 12.0.1, with the graphical display based on JavaFX. 
Typically, \textsc{TensorFlow} creates a computational graph that allows a just-in-time compiler to optimize its computations. Furthermore, this graph can be stored in a portable format and then run in other environments. However, this is not a useful approach for our prototype, since sufficient access to the internal architecture and operations of the model is not sufficient for the implementation of the method developed in this work.
The \texttt{tf.keras} module, on the other hand, provides a high-level API that makes it relatively easy to develop models. In the context of the prototype, however, it is particularly important that models trained with \textsc{Keras} can be saved and stored in external files in the \texttt{.h5}-format. This means that information about these models can basically be viewed outside of \textsc{TensorFlow}, which is essential for using the aforementioned models for demonstration purposes in the prototype. 

As described by the HDF Group in \cite{HDFGroup}, the Hierarchical Data Format 5, or HDF5 for short, is a concept for storing and managing complex and memory-intensive data. It consists of a data model, a file format, and libraries and applications for using the format in combination with the associated data model. The \texttt{.h5}-files, in which the information about the trained \textsc{Keras} models is stored externally, belong to this file format. 
The model architecture, which contains information about the individual neuron layers and their parameters, as well as the respective connections between these layers are stored in the \texttt{.h5}-file as \texttt{model\_config} in the form of a \textsc{JSON} string.
The values of the individual weights for each connection are stored within the \texttt{.h5}-file as lists grouped by the names of the associated layers.

\textsc{JavaScript} Object Notation, \textsc{JSON} \cite{JSONorg}, is a lightweight data exchange format in the form of a text format based on a subset of the \textsc{JavaScript} programming language, standard ECMA-262, third edition of December 1999, and defined in standard ECMA-404. The advantages of \textsc{JSON} are its ease of creation and readability for humans---which is central to us---and its ease of creation and parsing by machines. 
\textsc{JSON} uses ordered lists and name-value pairs to create a structure and to store objects, arrays, values, strings, and numbers in that structure.
The \texttt{MapperService} for reading information from \texttt{.h5}-files relies on the independently developed jHDF library by Mudd \cite{JHDF}.

\begin{figure*}[t]
\centerline{\includegraphics[width=.7\textwidth]{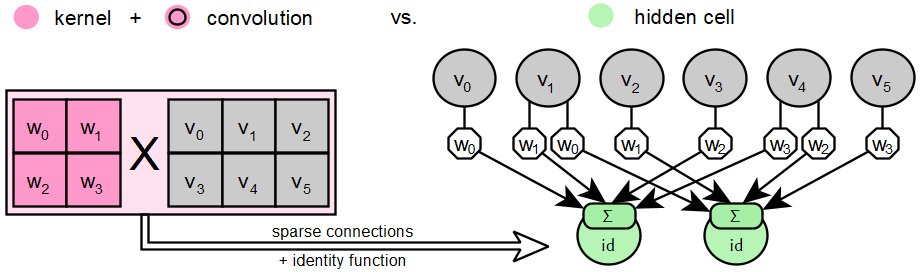}}
\caption{Description of the special features of a convolution using a kernel compared to hidden cells based on the descriptions in \cite{Goodfellow-et-al-2016}.}
\label{seide13}
\end{figure*}

\begin{figure*}[ht]
\centerline{\includegraphics[width=.7\textwidth]{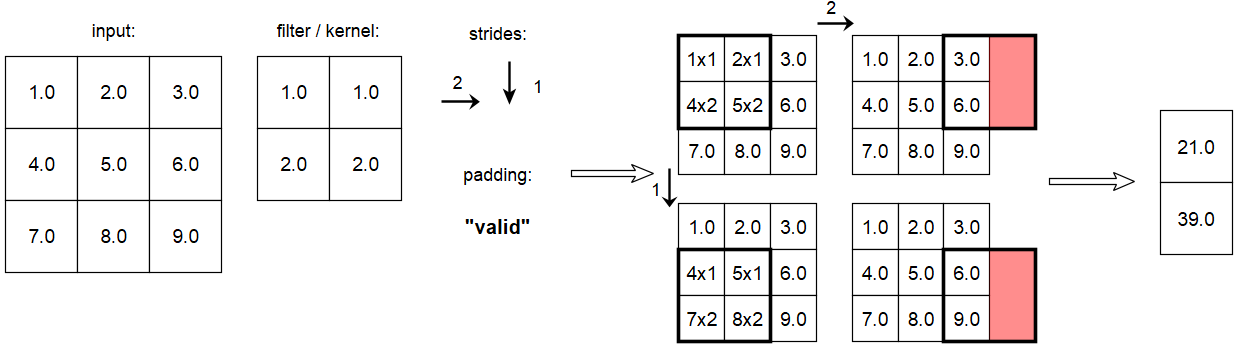}}
\caption{Example of a \texttt{Conv2D} layer processing a 2D input with given strides \texttt{[2,1]} and valid padding.}
\label{seide14}
\end{figure*}

\subsection{The layers}
In \textsc{Keras}, the \textit{input layer} is defined as an object of type \texttt{Input}, as described in \cite{KerasIO} under \textit{Layers API / Core layers / Input object}. This object represents the input function of an artificial neural network. This layer provides the values of the input vectors to the neural network in a defined form to be passed through the weighted input connections of the following layer.
\textsc{Keras} has no explicit output layers. Instead, the other available layer types serve as \textit{output layer} when instantiated as the last layer in a neural network. 
However, the last layer of a neural network is usually used in conjunction with a SoftMax function as the activation function. 
\textit{Feedforward layers} are implemented in \textsc{TensorFlow} with \textsc{Keras} using the \texttt{Dense} class, which already has a weight matrix for the input connections of its neurons via the \texttt{kernel} attribute. This weight matrix corresponds to a complete connection of the neurons to the previous layer, hence the name \texttt{Dense}, and can be used in the prototype to map the layer directly back to its feedforward form. 
The basic \textit{convolution layers} are provided in \textsc{Keras} with the classes \texttt{Conv1D}, \texttt{Conv2D} and \texttt{Conv3D}, which are based on the superclass \texttt{Conv}, whose convolution operation \texttt{\_convolutional\_op} uses the \textsc{TensorFlow} method \texttt{convolutional\_op}. \texttt{on\_v2} from the \texttt{tf.nn\_ops} package. This method is offered for use via the \texttt{tf.nn} package with the interface name \texttt{convolution}.
Therefore, for objects of the mentioned classes, a convolution kernel is stored in the \texttt{.h5}-file under \texttt{filters} and not an already completed weight matrix as it is the case for objects of the \texttt{Dense} class. 
It is also explained for the \texttt{tf.nn} module that the respective calculations of the weighted connections using the convolutional kernel depend on the applied zero padding, which is determined by the \texttt{padding} parameter \cite{TFAPI}. 
For further processing by the prototype, the convolutional layers are first converted from their special form into an equivalent feedforward form. This is always possible, since a convolutional layer is always functionally equivalent to a sparsely connected feedforward layer, as shown schematically in Fig.~\ref{seide13}. 
For a correct implementation, the additional attributes of the convolution, i.e., the padding and the steps used in each dimension, must also be included. An example of such a complete convolution according to \textsc{TensorFlow} is shown in Fig.~\ref{seide14}.

\begin{figure*}[t]
\centerline{\includegraphics[width=.9\textwidth]{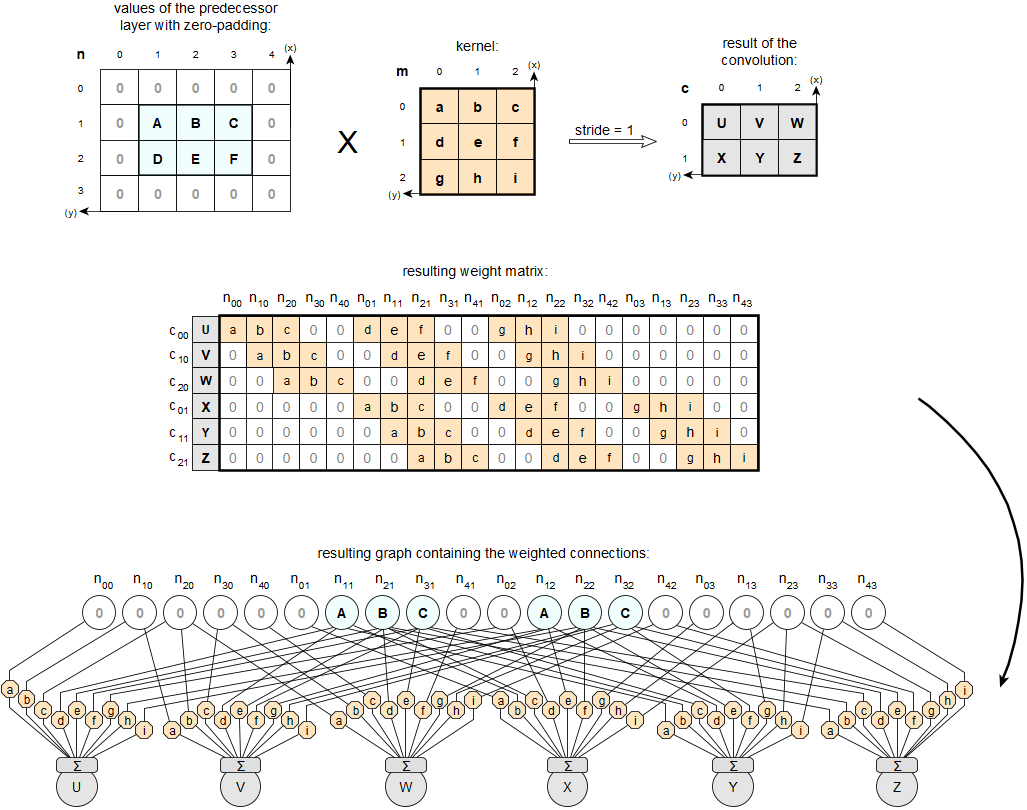}}
\caption{Example of the transfer of a 2D convolution into the filter of a feedforward layer using same padding.}
\label{seide15}
\end{figure*}

To convert the convolution kernel into a feedforward convolution layer, the convolution kernel is created as a separate object with stored functionality. The prototype can currently handle 1D, 2D, and 3D convolutions. 
Then the \texttt{Neuron} objects for the layer's \texttt{NeuralLayer} object are created by the corresponding \texttt{kernel} belonging to the appropriate subclass of \texttt{CNNKernel}.
After the \texttt{setNeurons} method of the \texttt{MapperService} has been fully processed for the \texttt{NeuralLayer} object of a layer, all filters and their \texttt{Neuron} objects with associated input weights are available in this object, mapping the feedforward form of the respective layer with all required functionalities. A schematic representation of the components involved in this procedure is shown in Fig.~\ref{seide15}, using a 2-dimensional example convolution, here with equal padding.

\subsection{The pooling}
In \textsc{Keras}, \textit{pooling} is implemented using classes for \texttt{MaxPooling}, \texttt{AveragePooling}, \texttt{Global}, \texttt{MaxPooling} and \texttt{GlobalAveragePooling}, where layer types for 1-dimensional, 2-dimensional and 3-dimensional input are available for these pooling types. Pooling is performed separately for each filter in a layer. 
For the prototype, only max-pooling and thus the classes \texttt{MaxPooling1D}, \texttt{MaxPooling2D} and \texttt{MaxPooling3D} are considered. Since the spatial processing of pooling is analogous to that of convolution, it is again possible to specify a padding parameter. 

\begin{figure}[ht]
\centerline{\includegraphics[width=\columnwidth]{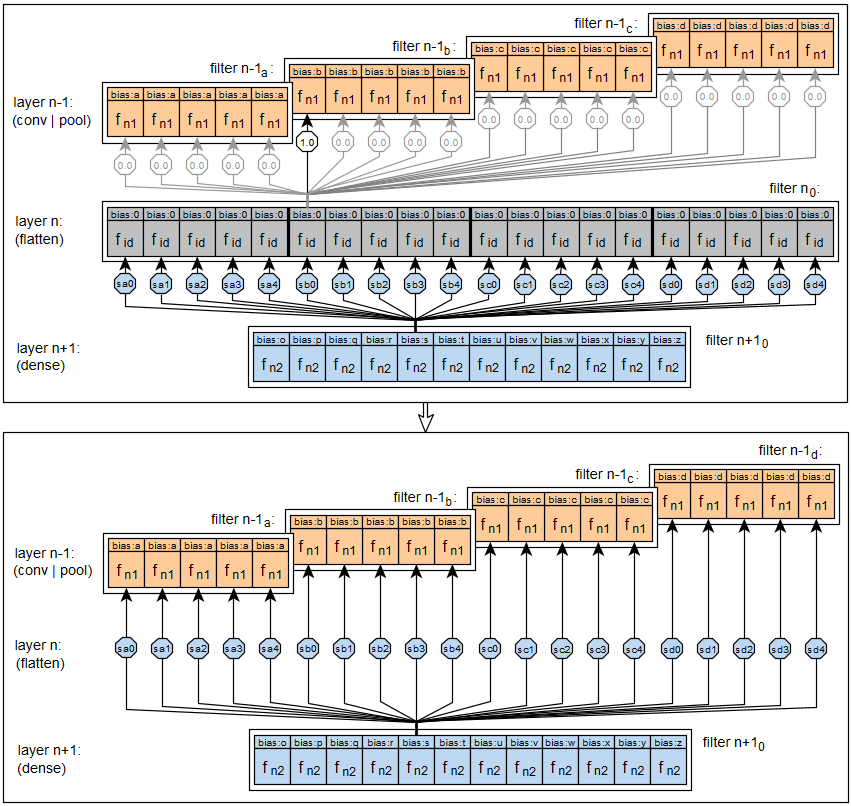}}
\caption{Visualization of the implementation of the principle of flattening layers in the prototype.}
\label{seide16}
\end{figure}

\textsc{TensorFlow} with \textsc{Keras} provides a number of layers that can be used to reshape the output of a previous layer to match a different, required input form of a subsequent layer. One such layer is the \textit{flattening layer}. This layer converts its respective input data, usually a multidimensional tensor containing the output activations of the filters of a previous layer, into a one-dimensional output tensor, which speeds up the process later on.
Fig.~\ref{seide16} shows the implementation of the flattening layer functionality in the prototype. The upper part of the figure shows the connection structure that would be needed if flat layers were interpreted as feedforward layers with neurons. The lower part shows how the connection structure from the upper part can be summarized.
As the figure shows, the prototype does not convert the flattening layers into a feedforward form, unlike the convolutional and pooling layers, although this would be possible. This is due to the function of the flattening layers, which only adjust the connections between their predecessor and successor layers. However, no new activation values are created or existing activation values are processed, so no new combinations are created by a flattening layer. Therefore the functionality of the flattening
layers in the prototype is not realized by neurons, but in the methods of the classes \texttt{NeuralNet} and \texttt{NeuralLayer}, which are relevant for the connections between the layers.

\subsection{The generation of hierarchical decision trees}
The class \texttt{TreeNode} represents the necessary data structure for the realization of node elements in a hierarchical decision tree. When generating hierarchical decision paths, it was not yet necessary to explicitly create such structures, since in all decision levels an edge $e$ is always followed by exactly one node $n$ and each node $n$ is followed by at most one edge $e$. So the vertices did not need to be considered explicitly. However, when combining multiple paths into a hierarchical decision tree, nodes are needed to carry information when creating the tree structure. 
Objects of type \texttt{Merger} each perform the procedure for combining their internal decision paths, recorded as \texttt{TreePath}, into a combined decision tree for an assigned \texttt{TreeEdge}. The result is the root node of the internal decision tree representing the assigned \texttt{TreeEdge}.

\begin{figure}[ht]
\centerline{\includegraphics[width=\columnwidth]{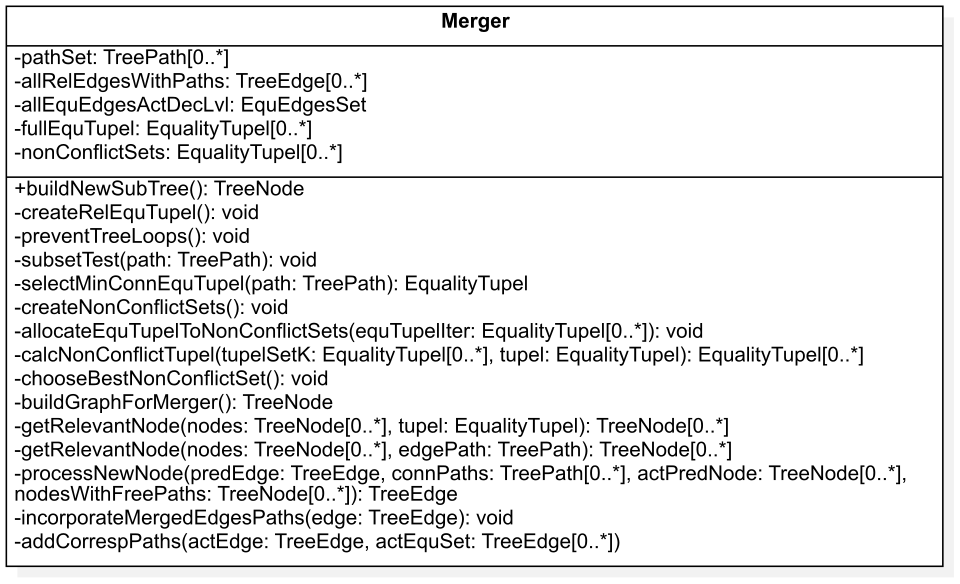}}
\caption{Representation of the \texttt{Merger} class, including its associated methods.}
\label{seide17}
\end{figure}

The main methods of the class \texttt{TreeNode} are shown in Fig.~\ref{seide17}.

\begin{figure}[ht]
\centerline{\includegraphics[width=\columnwidth]{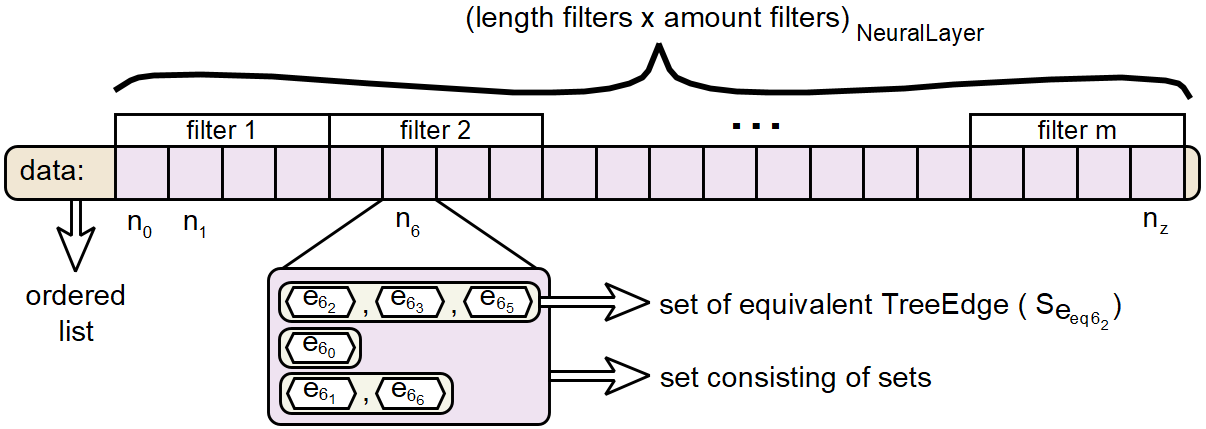}}
\caption{Visualization of the data structure behind the \texttt{data} attribute of the \texttt{EquEdgesSet} class.}
\label{seide18}
\end{figure}

A visualization of the \texttt{EquEdgesSet} class for the the \texttt{data} attribute is shown in Fig.~\ref{seide18}. This is an approach to perform the calculations for merging paths into a tree in the prototype without excessively inflating the heap.

\subsection{The graphical user interface}
\begin{figure*}[t]
\centerline{\includegraphics[width=\textwidth]{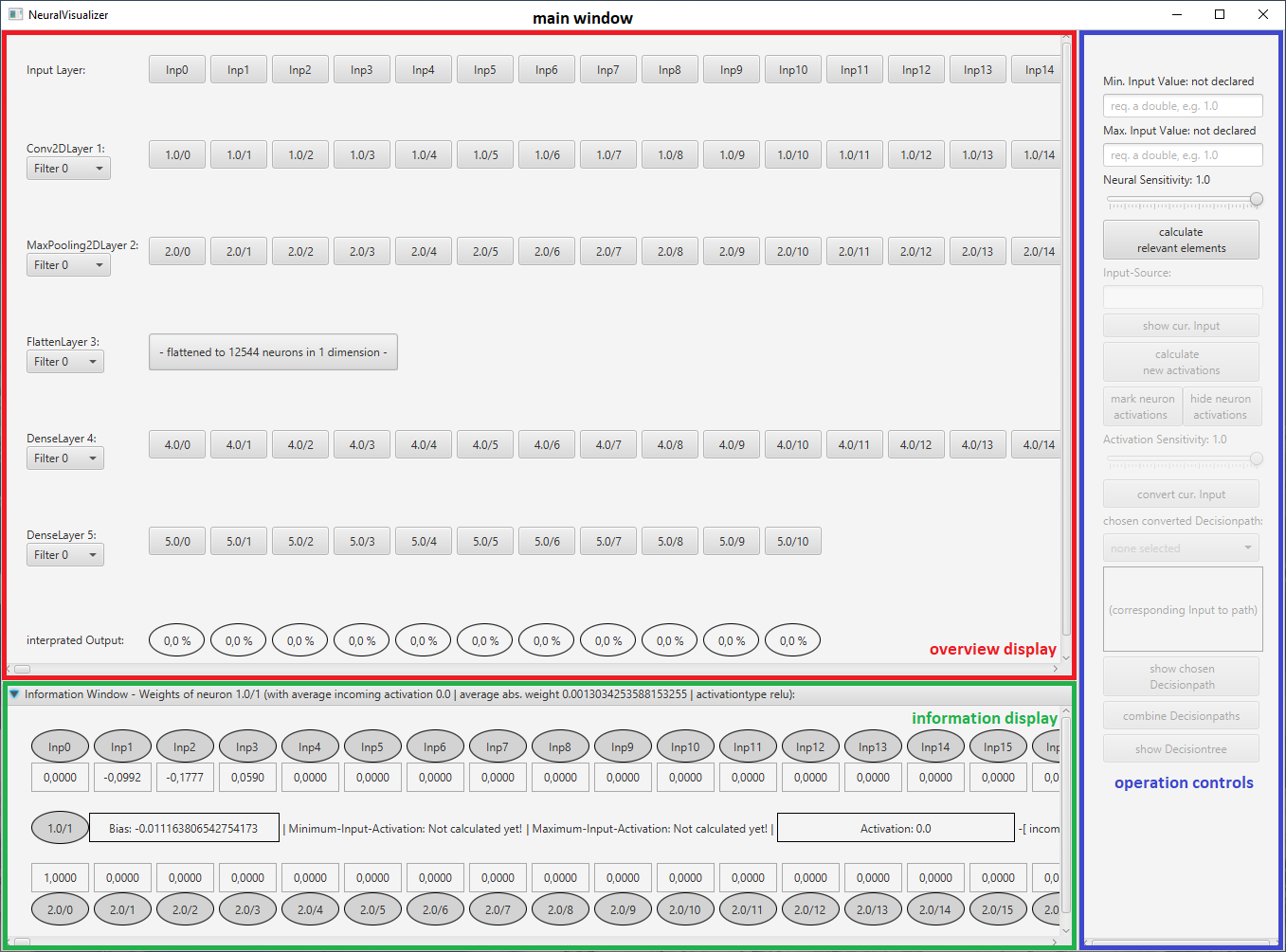}}
\caption{The start screen of the graphical user interface of the prototype after starting the program and generating the feedforward form of a read-in artificial neural network.}
\label{seide19}
\end{figure*}

The structure of the artificial neural network to be analyzed is loaded from the associated \texttt{.h5}-file by the main-class each time the prototype is started. After loading, the associated feedforward form is computed directly from the structure information of the neural network. 
This structure is presented in the overview display of its graphical user interface, allowing the user to initialize further process steps.
Fig.~\ref{seide19} shows the start screen featuring the main window. It is composed of three elements.

The \textit{overview display}, marked in red, shows an overview of the feedforward form of the analyzed neural network. The individual layers are listed from top to bottom, starting with the input layer and ending with the output layer. The numbering of each layer and a type description are shown on the left. Below this is a drop-down menu that can be used to change the display of the individual filters of a hidden layer, since only the neurons of one filter are displayed at a time. To the right, a neuron button with the name of the neuron is displayed for each of these neurons. For flattening layers, only the number of neurons to which the neurons from the filters of the previous layer are mapped is specified. For the output layer, the calculated output is shown. Clicking on a neuron button displays additional information about the corresponding neuron in the information display.

The \textit{information display}, marked in green, displays additional information about a selected neuron.
The header of the window displays the name of the selected neuron (\emph{layer number}, \emph{filter number}, \emph{neuron number}), the average value of the input activations, the average value of the weights of incoming connections and the associated activation function.
The selected neuron is displayed in the center. In addition, the associated bias value for the neuron's filter, the minimum and maximum possible output activation given the minimum and maximum possible input values for the input neurons, the calculated output activation of the neuron given the input activations, and the incoming activation sum are displayed.
Above and below the display of the selected neuron, the neurons of the associated filter of the predecessor and successor layer of this neuron are displayed, including the respective connection weights.

The \textit{operation controls}, marked in blue, provide input boxes for assigning different input values and buttons for initiating further processing steps.

\section{Summary and Conclusion}\label{sec5}
The following is a summary of the overall procedure to derive a hierarchical decision tree from a feedforward neural network and a set of input vectors given to the network.
The structural information of the analyzed network is read in and, if it is not already in this form, converted into a complete feedforward form. 
This is necessary, for example, for convolutional networks, which correspond to feedforward networks with a special connection structure in the convolutional, pooling, and flattening layers.
For the $n$ layers of the network, starting with the first layer, the respective estimates for the upper bound of the maximum output activation and the lower bound of the minimum output activation are determined. 
Then, the potentially relevant weighted connections and thus the potentially relevant neurons of the examined network are determined from the output layer to the input layer in order to reduce the computational effort.
For each of the input vectors to be examined, the following steps are performed:

\begin{enumerate}
\item the corresponding input vector is entered into the network and the resulting activations of all neurons for this input vector are calculated based on the minimized structure.
\item based on the calculated activations instead of the estimated boundaries, the relevant weighted connections and neurons of the minimized structure are calculated for the classification of the selected input vector.
\item based on the connections and neurons relevant for the classification of the selected input vector, the hierarchical decision path for said classification can now be derived. The associated symbol equivalents for this decision path are also created.
\end{enumerate}

The hierarchical decision paths that have been derived in the individual iterations are now combined to form a common hierarchical decision tree.

The procedure described above has been implemented and tested in a prototype. 
This proved that the desired derivation of equivalent symbol-based decision models from feedforward neural networks can be efficiently implemented. 
In this way, the inner workings of feedforward neural networks can be shown, the black-box character can be broken down, and the opaque decisions can be traced using the resulting decision trees.

For others to view, experiment with, build upon, and reuse, the code and test data for the prototype are available here
\cite{Seidel2025}.



\end{document}